\documentclass{article}

\usepackage{microtype}
\usepackage{graphicx}
\usepackage{subfigure}
\usepackage{booktabs} 

\usepackage{hyperref}

\usepackage[accepted]{icml2025}

\usepackage{amsmath}
\usepackage{amssymb}
\usepackage{mathtools}
\usepackage{amsthm}

\usepackage[capitalize,noabbrev]{cleveref}

\usepackage{xspace}

\makeatletter
\DeclareRobustCommand\onedot{\futurelet\@let@token\@onedot}
\def\@onedot{\ifx\@let@token.\else.\null\fi\xspace}

\def\eg{\emph{e.g}\onedot} 
\def\ie{\emph{i.e}\onedot}

\makeatother

\usepackage{paralist}
\usepackage[T1]{fontenc}
\newcommand*{\ourmethod}{\textsc{OptiAgent}\@\xspace}
\newcommand*{\ourdataset}{OptiDesignQA\@\xspace}

\newcommand{\PAR}[1]{\noindent{\bf #1}}

\usepackage{colortbl}
\usepackage{multirow}
\usepackage{arydshln}
\usepackage{caption}
\usepackage{makecell}

\theoremstyle{plain}

\theoremstyle{definition}

\theoremstyle{remark}

\usepackage[textsize=tiny]{todonotes}

\icmltitlerunning{\ourmethod: A Physics-Driven Agentic Framework for Automated Optical Design}

\begin{document}

\twocolumn[
\icmltitle{\ourmethod: A Physics-Driven Agentic Framework \\for Automated Optical Design}



\icmlsetsymbol{equal}{*}

\begin{icmlauthorlist}
\icmlauthor{Yuyu Geng}{equal,yyy}
\icmlauthor{Lei Sun}{equal,yyy,insait}
\icmlauthor{Yao Gao}{yyy}
\icmlauthor{Xinxin Hu}{comp}
\icmlauthor{Zhonghua Yi}{yyy}
\icmlauthor{Xiaolong Qian}{yyy}
\icmlauthor{Weijian Hu}{yyy}
\icmlauthor{Jian Bai}{yyy}
\icmlauthor{Kaiwei Wang}{yyy}
\end{icmlauthorlist}

\icmlaffiliation{yyy}{Zhejiang University}
\icmlaffiliation{insait}{INSAIT, Sofia University}
\icmlaffiliation{comp}{Xiaohongshu}


\icmlcorrespondingauthor{Lei Sun}{leo\_sun@zju.edu.cn}
\icmlcorrespondingauthor{Kaiwei Wang}{wangkaiwei@zju.edu.cn}

\icmlkeywords{Machine Learning, ICML}

]



\printAffiliationsAndNotice{\icmlEqualContribution} 

\begin{abstract}
Optical design is the process of configuring optical elements to precisely manipulate light for high-fidelity imaging. It is inherently a highly non-convex optimization problem that relies heavily on human heuristic expertise and domain-specific knowledge.
While Large Language Models (LLMs) possess extensive optical knowledge, their capabilities in leveraging the knowledge in designing lens system remain significantly constrained.
This work represents the first attempt to employ LLMs in the field of optical design. We bridge the expertise gap by enabling users without formal optical training to successfully develop functional lens systems.
Concretely, we curate a comprehensive dataset, named OptiDesignQA, which encompasses both classical lens systems sourced from standard optical textbooks and novel configurations generated by automated design algorithms for training and evaluation.
Furthermore, we inject domain-specific optical expertise into the LLM through a hybrid objective of full-system synthesis and lens completion. To align the model with optical principles, we employ Group Relative Policy Optimization Done Right (DrGRPO) guided by Optical Lexicographic Reward for physics-driven policy alignment. This reward system incorporates structural format rewards, physical feasibility rewards, light-manipulation accuracy, and LLM-based heuristics. Finally, our model integrates with specialized optical optimization routines for end-to-end fine-tuning and precision refinement.
We benchmark our proposed method against both traditional optimization-based automated design algorithms and LLM counterparts, and experimental results show the superiority of our method.
\end{abstract}

\begin{figure}
    \centering
    \includegraphics[width=0.98\linewidth]{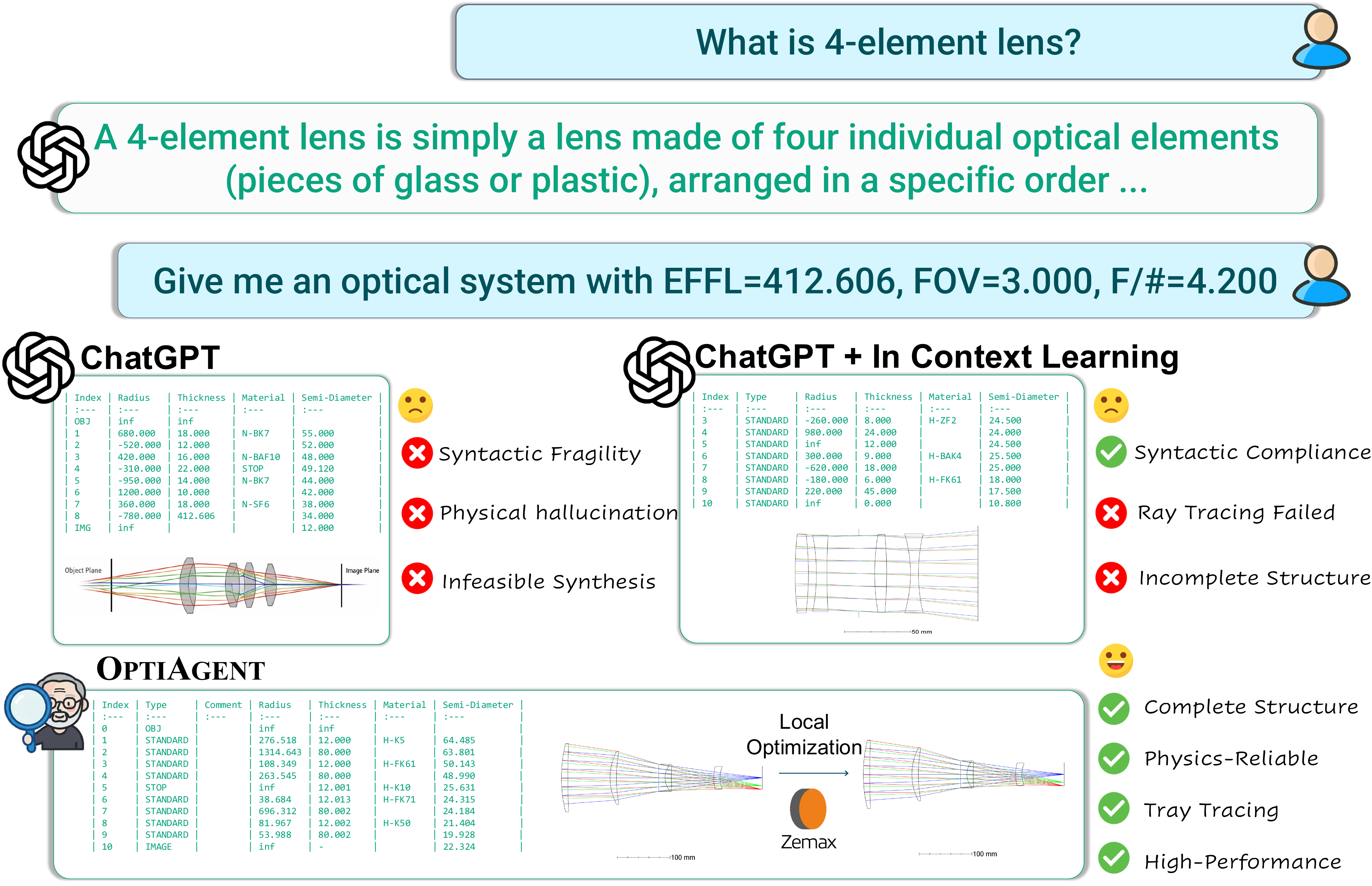}
    \caption{Despite possessing foundational optical knowledge, general-purpose LLMs (\textit{e.g.}, ChatGPT) fail to generate physically realizable systems, regardless of the prompting strategy. In contrast, \ourmethod bridges this gap by producing designs that strictly satisfy both practical and physical constraints.}
    \label{fig:teaser}
\end{figure}

\section{Introduction}
\label{sec:intro}

Optical lens design is a cornerstone of modern photonics, vital for technologies from smartphone imaging to advanced lithography~\cite{gao2022compact}. 
Fundamentally, it involves the precise spatial arrangement of refractive surfaces and materials to manipulate light propagation for specific imaging or non-imaging goals~\cite{gao2024design}.
The traditional workflow follows a two-stage paradigm: Human designers first rely heavily on tacit expertise, iterative experimentation, and intuition to propose a viable initial structure, which is then refined through high-dimensional, non-convex, and multi-objective optimization. 
This process remains labor-intensive and highly sensitive to initial configurations, necessitating continuous human intervention to manually adjust lens topologies and navigate complex trade-offs that automated optimizers often fail to resolve.

To address these challenges, various automated lens design methods have emerged, based on evolutionary algorithms (\eg QGSO~\cite{gao2025exploring}, NeuroOpti~\cite{gao2025neuro}). However, these approaches
often struggle with this discrete structural search, frequently producing physically invalid configurations or lacking diversity. Furthermore, these methods are highly time-consuming, often requiring dozens of days to converge, which limits the application in real-world scenarios.

\begin{figure*}
    \centering
    \includegraphics[width=0.98\linewidth]{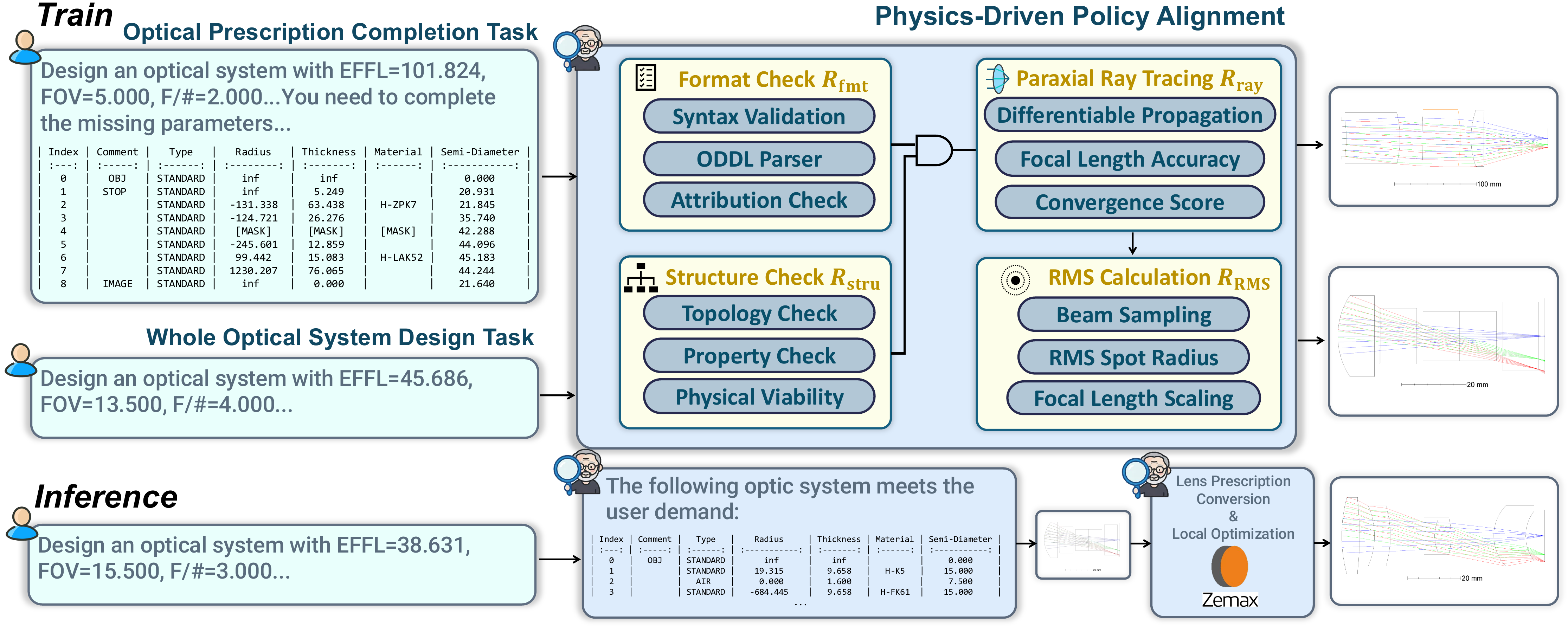}
    \caption{\textbf{Overview of the \ourmethod framework.}~Given natural language instruction, the system integrates both optical prescription completion task and whole optical system design task, a physics-driven policy with hierarchical rewards, and a function call of Zemax in a closed loop to synthesize optical prescriptions $\mathcal{L}$ from specifications $\mathcal{P}$. At inference, the generated initial structure $\mathcal{L}_0$ undergoes Zemax local optimization for final refinement.}
    \label{fig:main}
\end{figure*}

The whirlwind of progress in Large Language Models (LLMs)~\cite{devlin2019bert,chowdhery2023palm,lewis2020bart} has steadily transformed a myriad of specialized tasks, significantly boosting productivity by transcending simple text generation to embrace autonomous tool-use and complex problem-solving~\cite{nakano2021webgpt}.
Despite their vast internal knowledge of optical theories, we observe that LLMs often fail to translate textual understanding into viable design outputs. This discrepancy arises because optical design demands more than semantic recall. It requires navigating intricate inter-dependencies between geometric parameters and satisfying rigorous physical constraints. Conventional LLMs~\cite{touvron2023llama,openaigpto1,openaigpt4o,anthropic2024claude,deepseekai2025deepseekr1} lack the precision for high-fidelity structural optimization and often struggle with the ``spatial logic'' of lens systems.
As shown in Fig.~\ref{fig:teaser}, ChatGPT can answer questions in the field of optics, but it fails when given a specific optical design task because the optical system it provides lacks practical usability and manufacturability.

In this paper, we introduce \ourmethod, a pioneering agentic framework that reformulates optical lens design as a goal-oriented decision-making process within a physics-driven reinforcement learning (RL)~\cite{rafailov2023direct,shao2024deepseekmath}. We transform the LLM into a policy agent through three core innovations. First, we employ an Optical Prescription Completion task to inject ``physical intuition'' by forcing the agent to internalize geometric inter-dependencies. Second, we develop Optical Lexicographic Reward. This multi-level reward enhances the output form and physical consistency of LLM, while encouraging better optical system performance. It allows the agent to progressively align its policy with rigorous physical laws. Finally, the agent’s generated configurations serve as high-quality starting points for Zemax’s local optimization to achieve fine-grained, commercial-grade precision.
Compared with conventional LLMs in Fig.~\ref{fig:teaser}, \ourmethod is capable of generating high-fidelity optical systems that adhere to specified requirements and physical limitations.


At the data level, we introduce \ourdataset, the first dataset specifically curated for the post-training and evaluation of LLMs in optical lens design. To ensure both rigorous reliability and structural novelty, the dataset integrates classic architectures from authoritative textbooks with novel configurations generated by state-of-the-art automated global optimization algorithms~\cite{gao2025exploring}. This collection comprises a training set of 711 whole design tasks and 124 prescription completion tasks, alongside a dedicated test set of 80 whole optical system design tasks. Experiments demonstrate that \ourmethod significantly outperforms traditional optimizers and general-purpose LLMs in generating valid optical lens structures.

In summary, the main contributions of this work are:

\begin{compactitem}
    \item \textbf{Pioneering Agentic Framework:} We propose \ourmethod, the first agentic framework that reformulates optical lens design as a goal-oriented decision-making process, enabling general-purpose LLMs to solve complex optical design problems.
    
    \item \textbf{Optical Design Benchmark:} We curate \ourdataset, a tailored dataset comprising 711 textbook-based and algorithmically synthesized design tasks. It serves as the first dedicated resource for injecting physical intuition and evaluating LLMs in precision optics.
    
    \item \textbf{Physics-Driven Policy Alignment:} We introduce a Lexicographic Reward to enforce strict geometric and physical constraints. Experiments demonstrate that our method significantly outperforms existing LLMs in generating physically viable lens systems.
\end{compactitem}


\section{Related Work}
\label{sec:related_work}

\subsection{Automatic Optical Design}
Optical design, the systematic process of initializing and optimizing lens parameters to meet desired image quality under certain physical constraints, has been a topic of considerable importance in advancing how humans explore the unknown world~\cite{wang2015witnessing, gissibl2016sub, zhang2023large}. 
Since optical design is a daunting task, automatic optical design involving no or minimal human effort has always been the expectation of scientists, researchers, and optical engineers~\cite{yang2017automated}.
Modern optical design software, such as Zemax and CODE V, enables automated optimization workflows when provided with properly initialized configurations, significantly streamlining the lens design process. Recent advancements in computational methodologies have further reduced human intervention through diverse approaches, including heuristic search algorithms~\cite{guo2019new, zhang2020automated, gao2025exploring, liu2025global, gao2025neuro}, Deep Neural Networks~\cite{cote2019extrapolating, cote2021deep}, point-by-point optimization~\cite{yang2017automated, zhang2021towards}, and curriculum learning frameworks~\cite{yang2024curriculum}. 
These approaches predominantly rely on iterative optimization algorithms (\eg, gradient-based or evolutionary methods) that demand extensive computational resources and manual parameter tuning, resulting in design cycles comparable to manual optimization. While some deep learning methods have introduced retrieval-based frameworks for lens configuration matching, their inability to extrapolate beyond training data limits their applicability to novel specifications. In contrast, LLM-driven generative framework directly synthesizes physically plausible optical architectures from high-level specifications through physics-aware neural representations. This paradigm shift bypasses the limitations of retrieval-based systems by enabling open-ended design exploration within latent spaces.

\subsection{LLMs and LLM-agent}
Large Language Models (LLMs) have emerged as the cornerstone of Artificial General Intelligence (AGI). 
Beyond scaling laws, recent breakthroughs (\textit{e.g.}, LLaMA~\cite{touvron2023llama}, DeepSeek~\cite{liu2024deepseek}, Qwen3~\cite{yang2025qwen3} and GPT-5~\cite{singh2025openai}) underscore their potential as the cognitive engine for autonomous agents. 
By distilling exabytes of multi-domain knowledge into dense parameters, LLMs provide a robust prior for complex reasoning. 
Specifically, the integration of reinforcement learning~\cite{rafailov2023direct, shao2024deepseekmath} in recent reasoning models~\cite{guo2025deepseek} has pushed the boundaries of multi-step decision-making. 
To bridge the gap between general intelligence and practical utility, domain-specific LLM-agents~\cite{zhang2023xuanyuan,lai2025llmlight,cursor2024} have been developed in the areas of code generation, financial answering and traffic signal control.
Equipped with graph-based reasoning~\cite{besta2024graph} and dynamic tool utilization~\cite{qian2025toolrl}, LLM-guided workflow possesses the ability to solve complex problems.
Nevertheless, the current absence of specialized LLM or LLM-agent dedicated to the domain of optical design represents a formidable barrier, significantly impeding the progression toward intelligent automation within the optical engineering landscape.

\section{\ourmethod}
\label{sec:method}


\subsection{Preliminaries for Optics}
\label{subsec:pre}

A classical optical imaging system functions by mapping the radiance distribution from the Object Plane to the Image Plane through a coaxial arrangement of refractive elements. Central to this geometric transformation is the \textit{Effective Focal Length (EFFL)}, a Gaussian parameter that governs the system's fundamental scaling, optical power, and magnification. Beyond geometric scaling, regulating the luminous flux is equally critical, and this is achieved via the aperture stop, which limits the beam cross-section. The efficiency of this light-gathering capability is quantified by the \textit{F-number} (defined as the ratio of focal length to entrance pupil diameter), where a lower value implies greater light transmission and a shallower depth of field. Complementing these parameters, the spatial coverage of the system is described by the \textit{Field of View (FoV)}, which represents the maximum angular range of the incident scene. Ultimately, these optical properties must be physically realized within mechanical constraints: the system's overall compactness is constrained by the \textit{Total Track Length (TOTR)} (measured from the first lens vertex to the image sensor), while the necessary clearance for sensor packaging and filters is dictated by the \textit{Back Focal Length (BFL)}, defined as the physical distance from the posterior vertex of the last lens element to the Image Plane.
The details of the imaging process and underlying principles are elucidated in \textit{Supp.}~\ref{supp:fundamentals}.

\subsection{Motivation}
\label{subsec:motivation}
Despite possessing vast theoretical knowledge of optics, general-purpose LLMs consistently fail to produce functional lens designs. For example, while it can ostensibly define "What is a four-element optical system?", the specific design parameters it yields often lack practical utility and mechanical feasibility, rendering them unsuitable for real-world application. This ``capability gap'' stems from several fundamental challenges:

\PAR{Complexity of Physical Constraints:}~Optical design is governed by strict boundary conditions (\eg, non-negative edge thicknesses, no lens intersections). But LLMs that trained primarily on text lack the spatial reasoning to prevent physical violations that render a design unmanufacturable.

\PAR{Implicit Parameter Coupling:}~A lens system is not a collection of independent parameters; a change in a single surface's curvature necessitates reciprocal adjustments in thicknesses and glass types to maintain the focal length. LLMs struggle to internalize these non-linear inter-dependencies, leading to ``broken'' systems where light rays fail to converge.

\PAR{High-Precision Sensitivity:}~Unlike natural language where a typo is often negligible, a $1\%$ error in a lens's radius of curvature can lead to catastrophic aberrations. Text-based pre-training does not equip models with the numerical precision required for high-fidelity structural synthesis.

Therefore, we argue that optical design requires a \textit{physics-driven} approach rather than pure text generation. It is imperative to inject physical intuition into the model's weights and employ Reinforcement Learning (RL) to align its policy with rigorous optical constraints. This synergy enables the agent to generate robust, physically-grounded initial structures, which are subsequently refined via Zemax local optimization to achieve fine-grained, commercial-grade precision.

\subsection{\ourmethod Framework}
\label{subsec:overview}
Motivated by the aforementioned challenges, we propose \ourmethod, a physics-driven framework that systematically integrates physical intuition and environmental feedback. As illustrated in Fig~\ref{fig:main}, the framework operates as a closed-loop system consisting of three primary modules: the Policy Agent as an LLM-based decision-maker, the Physical Simulator for executing ray tracing and deriving raw physical metrics, and the Reward Evaluator for translating those metrics into hierarchical feedback.


Given a set of user-defined target specifications $\mathcal{P}$, which primarily constrains the system's scale and light-gathering capabilities through metrics such as EFFL ($f$), Field of View ($FoV$), and F-number ($F/\#$), the framework aims to synthesize a fully realized optical system. The objective is to generate a lens prescription $\mathcal{L}$ that is not only optically compliant but also physically viable—meaning it must possess positive element thicknesses, valid air gaps, and manufacturable glass selections. This generated prescription $\mathcal{L}$ serves as a comprehensive blueprint for the optical system, meticulously defining the sequential propagation path. It details the geometric profile (curvature and aperture) of every lens surface, assigns specific glass materials, and establishes the precise spatial configuration of critical planes, including the object plane, aperture stop, and the image plane.

At inference time, the optimized agent produces a robust initial structure $\mathcal{L}_0$, which is then refined through Zemax local optimization to achieve the final design. We restrict the use of Zemax’s fine-grained refinement to the inference stage because such high-precision optimization is a deterministic and computationally intensive process. This design ensures that the agent focuses on global structural synthesis during training while delegating high-precision numerical convergence to a specialized optical engine during deployment.

\begin{figure*}[t]
    \centering
    \includegraphics[width=0.98\linewidth]{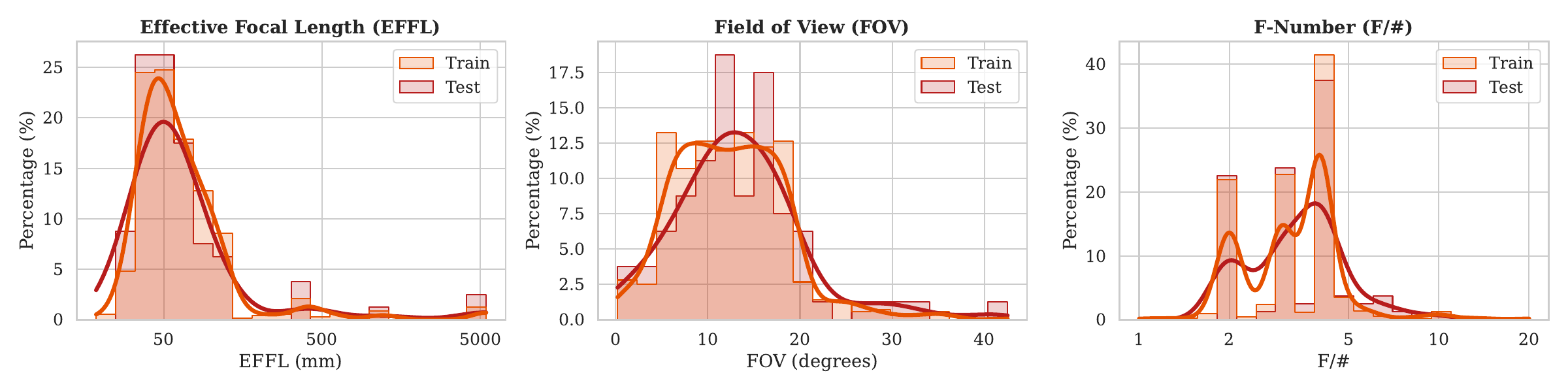}
    \vspace{-3pt}
    \caption{\textbf{Statistics of EFFL, FoV, and F-number in our \ourdataset dataset.} Our dataset covers a broad spectrum of specifications for commonly used optical lenses. Distinct peaks in F-number arises from the prevalence of standard engineering aperture settings in practical optical designs.}
    \label{fig:dataset}
    \vspace{-8pt}
\end{figure*}

\subsection{Knowledge Injection via Optical Prescription Completion} 
\label{subsec:knowledge_injection}





While LLMs excel at theoretical optics, their design capabilities are hindered by the scarcity of paired datasets linking specifications to lens structures. We introduce Optical Prescription Completion as an auxiliary task to inject domain-specific knowledge, forcing the model to internalize geometric inter-dependencies within lens systems in the RL training.

In this task, prompts for training consist of a design demand paired with a ``masked'' optical prescription. The agent must complete missing numerical values (\eg, radii, thicknesses) based on the remaining context:
This completion reward, combined with the subsequent Optical Lexicographic Reward, constitutes the objective function for our RL training pipeline.

\subsection{Physics-Driven Policy Alignment with Optical Lexicographic Reward}
\label{subsec:reward}
In application scenarios, \ourmethod is expected to return a feasible and complete optical lens design based solely on user demand $\mathcal{P}$. Consequently, the second half of our reward is derived from evaluating the validity and performance of the generated optical system using only $\mathcal{P}$ as input prompt. To this end, we propose the Optical Lexicographic Reward, which consists of a hierarchy of rewards ranging from encouraging proper output formatting and fundamental system parameters to physical feasibility and fulfillment of specific user demands. This hierarchical structure effectively drives the physics-driven RL process.

\PAR{Format Reward ($R_{\text{fmt}}$).}~Before evaluating any optical properties, the generated design must strictly adhere to the Optical Data Description Language (ODDL) format, including system specifications (\ie, EFFL and TOTR) and syntactically correct lens structure. A format score of 1 is given if all requisite elements can be parsed; otherwise, 0.

\PAR{Structure Reward ($R_{\text{stru}}$).}~Second, the lens structure must be physically sound, which is ensured via routine rule-based validation: (1) In a basic lens system, at least three surfaces should be included. The first and last surface must be defined as the object and image plane, respectively, with the aperture stop in between. (2) Every optical surface must include numerical values for curvature radius ($R$), thickness ($T$), and a valid material identifier (refractive and abbe). (3) The material of the penultimate surface must be designated as air, and its thickness (\ie, BFL) must be strictly positive ($BFL > 0$). Similarly, if any violation is detected, $R_{stru}$ is set to 0, and the entire evaluation terminates immediately. Otherwise, $S_{\text{stru}} = 1$.

\PAR{Paraxial Ray Tracing Reward ($R_{\text{ray}}$).}~Then, with a valid structure, we assess fundamental optical properties using first-order paraxial ray tracing to ensure the target EFFL is met, and the correct image plane positioning is maintained. More specifically, we implement a differentiable paraxial ray tracing engine to simulate light propagation. For each surface $i$, we update the ray's {angle} ($u$) and {height} ($y$) as follows:
\begin{align}
    &u'_i = \frac{n_i u_i - y_i (n'_i - n_i) c_i}{n'_i} \quad \text{(Bending the light)} \\
    &y_{i+1} = y_i + u'_i t_i \quad \quad \quad  \text{(Moving to next surface)}
    \label{eq:ray_tracing}
\end{align}
Here, $c$, $t$, and $n$ represent the surface curvature, thickness, and refractive index, respectively. By tracing a ray entering parallel to the optical axis, this recursion allows us to derive the calculated EFFL ($f_{\text{calc}}$), and the residual ray height at the image plane ($y_{\text{img}}$, ideally zero for perfect focus). Finally, we could derive the focal length accuracy score ($S_{\text{f}}$) and the convergence score ($S_{\text{c}}$) from Eq.~\ref{eq:ray_tracing}, and $R_{ray}$ is defined as the weighted sum of these two components. For the detailed derivation, kindly refer to the \textit{Supp}.


\PAR{RMS Reward ($R_{\text{RMS}}$)}~To enforce uniform image quality, we optimize for the worst-case scenario. The $R_{\text{RMS}}$ is calculated as:\begin{equation}R_{\text{RMS}} = \exp\left( -\frac{\max_{k} \sigma_k}{\gamma} \right)\end{equation}where $\sigma_k$ is the RMS spot radius for field $k$. The scaling factor $\gamma = \max(0.05, 0.01 \cdot f_{\text{effl}})$ dynamically adapts to the system's scale, tolerating a spot size relative to the focal length. For a detailed derivation of the beam sampling strategy, kindly refer to the \textit{Supp.}

\PAR{Total Lexicographic Reward.}~We aggregate these components into a hierarchical reward $R$. To ensure efficiency, $R_{\text{ray}}$ is computed only if $R_{\text{fmt}}$ and $R_{\text{stru}}$ are satisfied. Furthermore, the high-order $R_{\text{RMS}}$ is activated via a gating indicator $\delta_{\text{pass}}$ only when first-order properties (EFFL and $y_{\text{img}}$) meet specific convergence thresholds. The total reward is defined as:
\begin{equation}
    R_{\text{lex}} = R_{\text{fmt}} \cdot R_{\text{stru}} \cdot (R_{\text{ray}} + \delta_{\text{pass}} R_{\text{RMS}})
\end{equation}
Detailed gating criteria and the RL training process using the DrGRPO strategy are provided in the \textit{Supp.}

\section{\ourdataset Dataset}
\label{sec:dataset}





\begin{figure*}[t]
    \centering
    \includegraphics[width=0.98\linewidth]{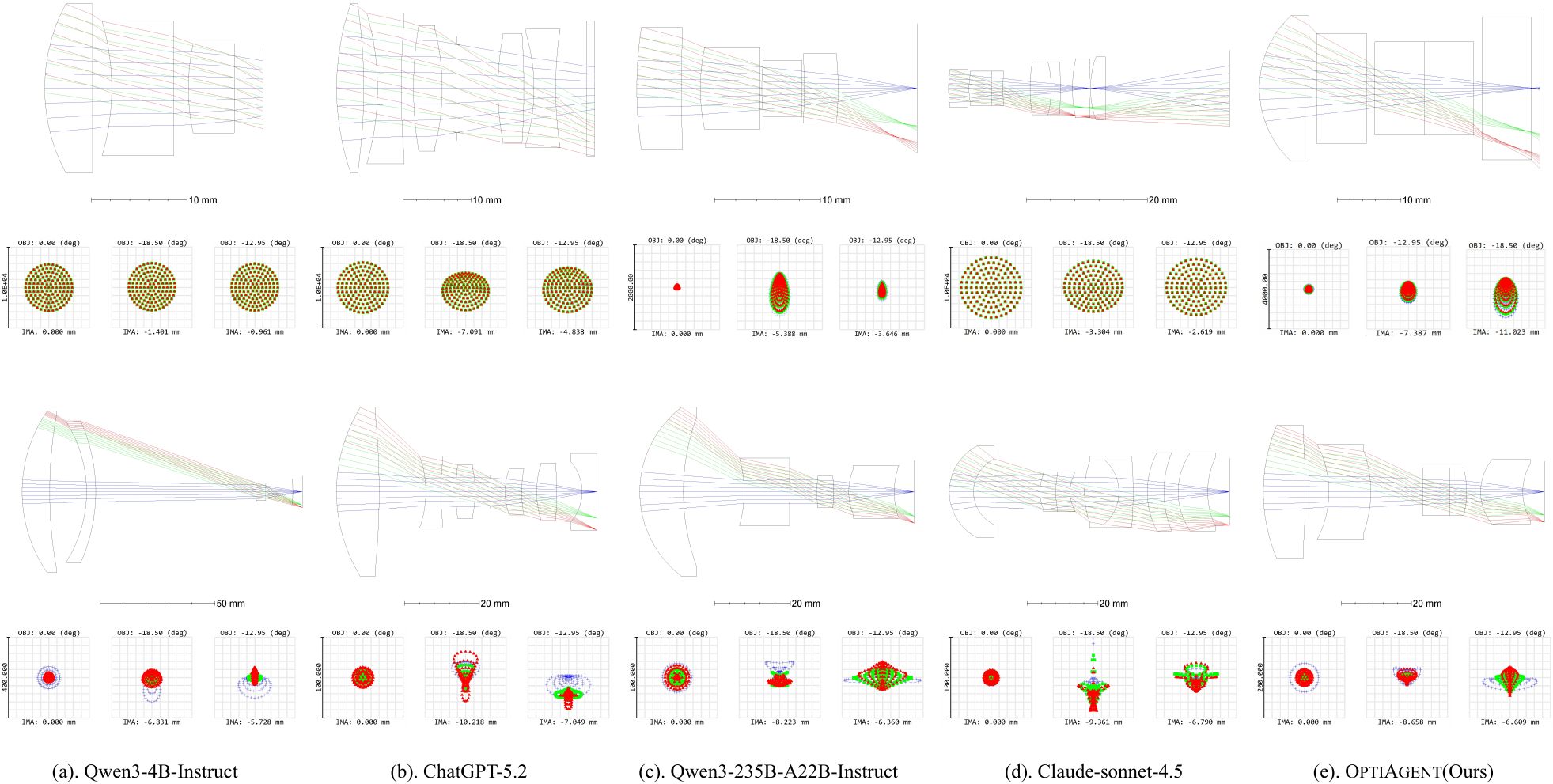}
    \caption{\textbf{Layouts of representative designs from all evaluated methods.} For each optical system, the RMS spot diagrams for three distinct FoVs are displayed. The first row illustrates the initial layouts produced by each method before fine-grained optimization in Zemax, while the second row shows the corresponding results after optimization. Ray colors blue, green, and red represent wavelengths of 0.485 $\mu$m, 0.588 $\mu$m, and 0.656 $\mu$m, respectively. Even when compared against significantly larger architectures like Qwen3-235B, our method utilizing Qwen3-4B as a foundation model exhibits better performance.}
    \vspace{-8pt}
    \label{fig:results}
\end{figure*}

\begin{table*}[t]
\scriptsize 
\newcommand{\best}[1]{\textbf{#1}}
\centering
\setlength{\tabcolsep}{3mm} 
\renewcommand{\arraystretch}{1.25} 
\resizebox{0.9\textwidth}{!}{
\begin{tabular}{l|c|cc|cc|cc} 
\toprule
\multirow{2}{*}{\textbf{Model}} & 
\multirow{2}{*}{\textbf{SR(\%)} $\uparrow$} & 
\multicolumn{2}{c|}{\textbf{EFFL Relative Error} $\downarrow$} & 
\multicolumn{2}{c|}{\textbf{Initial RMS} $\downarrow$} & 
\multicolumn{2}{c}{\textbf{Final RMS} $\downarrow$} \\
\cmidrule(lr){3-4} \cmidrule(lr){5-6} \cmidrule(lr){7-8}
 & & \textbf{Avg(\%)} & \textbf{Std} & \textbf{Avg(um)} & \textbf{Std} & \textbf{Avg(um)} & \textbf{Std} \\ 
\midrule

Qwen3-4B-Instruct & 
56.9 & 
42.1 & 
0.29 & 
12405.19 & 
7578.91 & 
758.30 & 
227.46 \\

ChatGPT-5.2 & 
72.0 & 
37.2 & 
0.22 & 
9942.35 & 
14033.07 & 
249.64 & 
172.65 \\

Qwen3-235B-22B-Instruct & 
77.8 & 
54.7 & 
0.19 & 
3288.30 & 
2107.08 & 
127.35 & 
176.32 \\

Claude-Sonnet-4.5 & 
81.8 & 
35.3 & 
0.22 & 
5847.89 & 
6113.31 & 
155.02 & 
164.47 \\

\textbf{\ourmethod (Ours)} & 
\best{90.1} & 
\best{1.0} & 
\best{$1.06\times10^{-3}$} & 
\best{672.09} & 
\best{138.10} & 
\best{41.23} & 
\best{76.14} \\

\bottomrule
\end{tabular}
}
\vspace{-3pt}

\caption{\textbf{Performance comparison with LLMs} on our \ourdataset test set. Best results in \best{bold}.}
\label{tab:comparison_llms}
\vspace{-12pt}

\end{table*}

\newcommand{\best}[1]{\textbf{#1}}
\newcommand{\second}[1]{\textcolor{blue}{\textbf{#1}}}
\newcommand{\redval}[1]{\textcolor{red}{#1}}

\begin{table*}[t]
\scriptsize 
\centering
\resizebox{0.9\textwidth}{!}{
\setlength{\tabcolsep}{3mm} 
\renewcommand{\arraystretch}{1.25} 

\begin{tabular}{l|c|cc|cc|cc} 
\toprule
\multirow{2}{*}{\textbf{Reward Combination}} & 
\multirow{2}{*}{\textbf{SR(\%)} $\uparrow$} & 
\multicolumn{2}{c|}{\textbf{EFFL Relative Error} $\downarrow$} & 
\multicolumn{2}{c|}{\textbf{Initial RMS} $\downarrow$} & 
\multicolumn{2}{c}{\textbf{Final RMS} $\downarrow$} \\
\cmidrule(lr){3-4} \cmidrule(lr){5-6} \cmidrule(lr){7-8}
 & & \textbf{Avg(\%)} & \textbf{Std} & \textbf{Avg(um)} & \textbf{Std} & \textbf{Avg(um)} & \textbf{Std} \\ 
\midrule

Ray Tracing & 
95.3 & 
1.9 & 
$3.80\times 10^{-3}$ & 
717.4 & 
96.3 & 
1422.7 & 
1449.3 \\

Ray Tracing + RMS & 
0.00 & 
/ & 
/ & 
/ & 
/ & 
/ & 
/ \\

Ray Tracing + RMS + LLM-as-judge & 
83.7 & 
28.2 & 
$2.06\times 10^{-17}$ & 
621.3 & 
$5.01 \times 10^{-14}$ & 
305.1 & 
$1.25\times 10^{-14}$ \\

Ray Tracing $\rightarrow$ LLM-as-judge & 
42.5 & 
1.2 & 
\best{$8.00\times 10^{-4}$} & 
4574.0 & 
1711.8 & 
344.9 & 
222.18 \\

Ray Tracing $\rightarrow$ LLM-as-judge $\rightarrow$ RMS & 
\best{93.7} & 
1.6 & 
$2.40\times 10^{-3}$ & 
1809.2 & 
1112.6 & 
127.3 & 
123.9 \\

Ray Tracing $\rightarrow$ (RMS + LLM-as-judge) & 
70.7 & 
1.4 & 
$2.20\times 10^{-3}$ & 
1484.5 & 
352.2 & 
101.2 & 
129.1 \\

(Ray Tracing + LLM-as-judge ) $\rightarrow$ RMS & 
90.6 & 
1.3 & 
$4.80\times 10^{-3}$ & 
1004.8 & 
\best{74.1} & 
89.7 & 
140.1 \\

Ray Tracing $\rightarrow$ RMS (\textbf{Ours}) & 
90.1 & 
\best{1.0} & 
$1.10\times 10^{-3}$ & 
\best{672.1} & 
138.1 & 
\best{41.2} & 
\best{76.1} \\

\bottomrule
\end{tabular}
}
\vspace{-3pt}

\caption{\textbf{Ablation study on the design of reward function.} ``+'' means combining and ``$\rightarrow$" signifies a threshold-based activation where the next stage is triggered only after the current criterion is satisfied.}
\label{tab:ablation_reward_v2}
\vspace{-12pt}
\end{table*}

We introduce \ourdataset, the first dataset specifically curated for the post-training and evaluation of LLMs in optical lens design. The training set consists of $711$ pairs of whole optical design tasks and $124$ optical prescription completion tasks. For evaluation, the test set includes $80$ whole optical system design tasks. To ensure both rigorous reliability and structural novelty, the dataset consists of both classic lens architectures sourced from authoritative textbooks, and novel configurations generated by state-of-the-art (SOTA) automated global optimization algorithms~\cite{gao2025exploring}. This dual-source approach provides a comprehensive benchmark that spans from fundamental expert knowledge to novel design spaces.

Fig.~\ref{fig:dataset} shows the statistics of EFFL, FoV, and F-number in our \ourdataset dataset. Our dataset features comprehensive coverage of the parameter space characteristic of standard and novel optical lens designs. Note that the F-number distribution exhibits distinct peaks rather than a continuous spread. This is attributed to the prevalence of industry-standard aperture settings (\eg, $F/2.0$, $F/2.8$, $F/4.0$) within our dataset. Such a distribution reflects the practical design requirements of commonly used optical systems.

\section{Experiments}
\label{sec:experiments}
\subsection{Evaluation Protocol.}~To ensure a statistically rigorous evaluation, we assess each method across several design specifications. For all the methods, to ensure the robustness and minimize the influence of random noise, we generate $10$ candidate prescriptions per specification. The evaluation metrics are prioritized in the following hierarchical order of importance:

\PAR{Success Rate (SR).}~Defined as the proportion of generated designs that are physically viable (\eg, no surface intersections) and can be successfully parsed by the optical engine for paraxial evaluation.

\PAR{EFFL Relative Error.~}Calculated as the percentage error between the generated EFFL and the target specification. We only calculate the EFFL Relative Error prior to Zemax's local optimization. This is because optimization can enforce the target EFFL at the expense of physical plausibility, potentially leading to structural breakdown. Consequently, comparing post-optimization EFFL values would be statistically insignificant.

\PAR{Root Mean Square (RMS).}~We use Root Mean Square (RMS) spot radius to evaluate the optical quality. The RMS spot radius quantifies the average spread of rays from the ideal image point, providing a statistical measure of the system's geometric aberration. More specifically, we evaluate the RMS of optical systems from both before and after Zemax's local optimization. which are denoted as ``Initial RMS'' and ``Final RMS'', respectively.

\PAR{Statistical Robustness.}~To account for the stochastic nature of generative models, we report both the average metrics (Avg) and the standard deviation (Std) of the percentage errors and RMS values across the $10$ samples, which reflects the consistency of the model's physical reasoning.

For any practical optical system, meeting the target specifications is the fundamental requirement, while the RMS spot radius serves as a secondary optimization goal. In our experimental setup, the FoV and F-number are strictly enforced at the initialization stage, resulting in no measurable error. Therefore, we prioritize the EFFL Relative Error as the decisive criterion, as the analysis of RMS becomes physically meaningful only after the target focal length has been successfully achieved.

\begin{table*}[t]
\scriptsize 
\centering
\setlength{\tabcolsep}{4mm} 
\renewcommand{\arraystretch}{1.25} 
\resizebox{0.9\textwidth}{!}{
\begin{tabular}{l|c|cc|cc|cc} 
\toprule
\multirow{2}{*}{\textbf{Ratio of the masked data}} & 
\multirow{2}{*}{\textbf{SR(\%)} $\uparrow$} & 
\multicolumn{2}{c|}{\textbf{EFFL Relative Error} $\downarrow$} & 
\multicolumn{2}{c|}{\textbf{Initial RMS} $\downarrow$} & 
\multicolumn{2}{c}{\textbf{Final RMS} $\downarrow$} \\
\cmidrule(lr){3-4} \cmidrule(lr){5-6} \cmidrule(lr){7-8}
 & & \textbf{Avg(\%)} & \textbf{Std} & \textbf{Avg(um)} & \textbf{Std} & \textbf{Avg(um)} & \textbf{Std} \\ 
\midrule

0 & 
28.8 & 
41.3 & 
$2.17 \times 10^{-17}$ & 
268.1 & 
$1.24 \times 10^{-14}$& 
601.8 & 
$5.50 \times 10^{-14}$\\

$50\%$ & 
64.3 & 
27.5 & 
$2.06 \times 10^{-17}$ & 
637.4 & 
$4.91 \times 10^{-14}$& 
256.4 & 
31.2 \\

$100\%$ & 
90.1 & 
9.8 & 
0.001 & 
672.1 & 
138.1 & 
41.23 & 
76.14 \\

\bottomrule
\end{tabular}
}
\vspace{-6pt}

\caption{\textbf{Ablation study on the Optical Prescription Completion.} Different ratios of masked data are tested.}
\vspace{-8pt}

\label{tab:ablation_completion}
\end{table*}

\begin{table*}[t]
\scriptsize 
\centering
\setlength{\tabcolsep}{5mm} 
\renewcommand{\arraystretch}{1.25} 
\resizebox{0.9\textwidth}{!}{
\begin{tabular}{l|c|cc|cc|cc} 
\toprule
\multirow{2}{*}{\textbf{Train Method}} & 
\multirow{2}{*}{\textbf{SR(\%)} $\uparrow$} & 
\multicolumn{2}{c|}{\textbf{EFFL Relative Error} $\downarrow$} & 
\multicolumn{2}{c|}{\textbf{Initial RMS} $\downarrow$} & 
\multicolumn{2}{c}{\textbf{Final RMS} $\downarrow$} \\
\cmidrule(lr){3-4} \cmidrule(lr){5-6} \cmidrule(lr){7-8}
 & & \textbf{Avg~(\%)} & \textbf{Std} & \textbf{Avg} & \textbf{Std} & \textbf{Avg} & \textbf{Std} \\ 
\midrule

SFT $\to$ RL & 
85.0 & 
$28.5\%$ & 
\best{0.0003} & 
851.7 & 
\best{0.26} & 
200.9 & 
\best{20.9} \\

SFT & 
36.1 & 
$52.8\%$ & 
0.27 & 
5367.2 & 
4083.1 & 
123.84 & 
244.18 \\

RL (\textbf{Ours}) & 
\best{90.1} & 
\best{$0.98\%$} & 
0.001 & 
\best{672.1} & 
138.1 & 
\best{41.23} & 
76.14 \\

\bottomrule
\end{tabular}
}
\vspace{-3pt}

\caption{\textbf{Ablation on training paradigm.} Comparison between SFT, SFT followed by RL, and direct RL (Ours). \textbf{Bold} indicates the best performance.}
\vspace{-12pt}

\label{tab:ablation_paradigm}
\end{table*}

\subsection{Experimental Setup}
\label{subsec:setup}
\PAR{Baseline Methods.}~Given the absence of directly comparable frameworks for LLM-based optical design, we evaluate \ourmethod against a selection of representative baseline language models with in-context learning~\cite{dong2024survey}. We exclude comparisons with optimization-based methods, such as QGSO, due to their inherent computational constraints. Generating a single lens design under specific requirements typically necessitates several days of computation, which is inconsistent with our objective of providing near-instantaneous optical design generation. Furthermore, the scope of our work is to democratize optical design for users without specialized optical expertise. Consequently, the ability to generate prescriptions directly through natural language rather than through complex technical parameter tuning is of paramount importance to our framework. For large language models, we include both the state-of-the-art closed-source model ChatGPT-5.2~\cite{singh2025openai}, Claude Sonnet 4.5, and the leading open-source model Qwen3 (4B \& 235B)~\cite{yang2025qwen3}. Note that all the competitors utilize in-context-learning with few-shot optical design exemplars for fair comparison.


\PAR{Implementation Details.}~The training of \ourmethod is implemented using the verl framework, with the policy optimized via the DrGRPO~\cite{chen2025dra} algorithm to handle high-dimensional optical design reasoning. During the reinforcement learning stage, we perform a rollout of $G=16$ candidate responses for each prompt to compute the group-relative advantage. The training process uses a batch size of $2$, with the KL-divergence coefficient $\beta$ set to $0.01$ and the entropy weight maintained at $0.001$. To accommodate the complexity of long-sequence lens prescriptions, the maximum prompt length and maximum new tokens are configured to $3072$ and $7168$, respectively. For computational efficiency, we employ the vLLM engine for accelerated rollout with a GPU memory utilization of $0.2$, allowing for optimal resource allocation on NVIDIA H100 GPUs. The model is trained until the policy reaches empirical convergence across our validation benchmarks.





\subsection{Main Results}
\label{subsec:results}
Tab.~\ref{tab:comparison_llms} presents the final results of all evaluated methods on our \ourdataset. Regarding the most critical metric, Success Rate (SR), our method achieves state-of-the-art performance with a value exceeding $95\%$. This result underscores the physical robustness of our approach and validates the effectiveness of the proposed Physics-Driven Policy Alignment.

For the EFFL Relative Error, our method achieves a marginal $1\%$ error. This represents a substantial leap in precision compared to all baselines and demonstrates superior instruction-following capabilities. Furthermore, our approach yields the most favorable pre-optimization RMS spot radius, outperforming competitors by an order of magnitude. These results are primarily attributed to the explicit integration of physically rigorous ray tracing and RMS calculations during the RL training process.

While some baselines may exhibit smaller post-optimization RMS values, this is primarily attributed to their failure to adhere to the target EFFL. In cases where the EFFL error is excessively large, any analysis of the RMS metric loses its physical significance because it no longer reflects a valid design tailored to the user's specific requirements.

Fig.~\ref{fig:results} visualizes the lens designs generated by each evaluated method. We observe that our \ourmethod exhibits impressive optical performance even prior to the fine-grained optimization in Zemax, a capability that other baseline models lack. Only the Qwen3-235B model, which is two orders of magnitude larger, demonstrates comparable initial results. This underscores that specialized physical alignment outweighs model scaling for complex optical design.

\subsection{Ablation Study}
\label{subsec:ablation}
Here we do a comprehensive ablation study on our \ourmethod, mainly about the design of the reward function, the effectiveness of the Optical Prescription Completion task, and the training paradigm.

\PAR{Design of Reward Function.} Tab.~\ref{tab:ablation_reward_v2} shows the results. Adopting the ray tracing only achieves a SR of $90\%$, showing the effectiveness of the physics-driven policy. 
The naive combination of ray tracing and RMS rewards leads to training divergence. This instability occurs because introducing RMS constraints before achieving basic ray tracing validity creates conflicting optimization objectives. Our hierarchical approach avoids this issue by ensuring the model satisfies fundamental structural requirements before initiating fine-grained RMS optimization. This outcome validates our threshold-triggered strategy, which facilitates a stable transition from basic structural formation to precision refinement. More details about the details refer to \textit{supp.} The introduction of LLM-as-a-judge does not yield significant performance gains. This marginal impact is expected, as large language models lack the precision required for rigorous optical calculations, which aligns with our motivation.

\PAR{Optical Prescription Completion task.}~Tab.~\ref{tab:ablation_completion} shows that the introduction of the Optical Prescription Completion task improves performance, validating the importance of knowledge injection. However, we also observe that if the masking ratio exceeds $100\%$, the training process collapses.

\PAR{Training Paradigm.} We also evaluate Supervised Fine-Tuning (SFT) for post-training, as detailed in Tab.~\ref{tab:ablation_paradigm}, using a learning rate of $1 \times 10^{-5}$ for 3 epochs. In contrast, our RL training utilizes a lower learning rate of $1 \times 10^{-6}$ for a single epoch. In our case, SFT actually leads to a decrease in Success Rate (SR). This occurs because optical design demands robust physical reasoning capabilities, which are better cultivated through policy alignment rather than simple pattern imitation.

\section{Conclusion}
\label{sec:conclusion}

We presented \ourmethod, a pioneering agentic framework that bridges the gap between Large Language Models and the rigorous physical constraints of optical lens design. To address the inherent "spatial logic" limitations of standard LLMs, we introduced a novel training paradigm combining an Optical Prescription Completion task with a physics-driven Hierarchical Group Relative Policy Optimization (DrGRPO). This approach enables the agent to internalize geometric inter-dependencies and progressively align with optical laws. We also released \ourdataset, a diverse benchmark to facilitate standardized evaluation in this domain. Extensive experiments demonstrate that \ourmethod significantly outperforms traditional evolutionary algorithms and general-purpose LLMs, producing high-quality initial structures ready for professional optimization in Zemax. This work showcases the potential of LLMs for complex physical problems and paves the way for autonomous optical engineering systems.

\bibliography{main}

@String(AAAI = {AAAI})

@article{chen2025dra,
  title={Dra-grpo: Exploring diversity-aware reward adjustment for r1-zero-like training of large language models},
  author={Chen, Xiwen and Zhu, Wenhui and Qiu, Peijie and Dong, Xuanzhao and Wang, Hao and Wu, Haiyu and Li, Huayu and Sotiras, Aristeidis and Wang, Yalin and Razi, Abolfazl},
  journal={arXiv preprint arXiv:2505.09655},
  year={2025}
}

@misc{shao2024deepseekmath,
      title={DeepSeekMath: Pushing the Limits of Mathematical Reasoning in Open Language Models}, 
      author={Zhihong Shao and Peiyi Wang and Qihao Zhu and Runxin Xu and Junxiao Song and Xiao Bi and Haowei Zhang and Mingchuan Zhang and Y. K. Li and Y. Wu and Daya Guo},
      year={2024},
      eprint={2402.03300},
      archivePrefix={arXiv},
      primaryClass={cs.CL},
      url={https://arxiv.org/abs/2402.03300}, 
}

@article{gao2022compact,
  title={Compact and lightweight panoramic annular lens for computer vision tasks},
  author={Gao, Shaohua and Sun, Lei and Jiang, Qi and Shi, Hao and Wang, Jia and Wang, Kaiwei and Bai, Jian},
  journal={Optics Express},
  volume={30},
  number={17},
  pages={29940--29956},
  year={2022},
  publisher={Optica Publishing Group}
}

@article{guo2019new,
  title={New automatic optical design method based on combination of particle swarm optimization and least squares},
  author={Guo, Dabo and Yin, Liang and Yuan, Guang},
  journal={Optics Express},
  volume={27},
  number={12},
  pages={17027--17040},
  year={2019},
  publisher={Optica Publishing Group}
}

@inproceedings{zhang2020automated,
  title={Automated design of machine vision lens based on the combination of particle swarm optimization and damped least squares},
  author={Zhang, Jiajun and Cen, Zhaofeng and Li, Xiaotong},
  booktitle={Optical Design and Testing X},
  volume={11548},
  pages={261--272},
  year={2020},
  organization={SPIE}
}

@article{yang2024curriculum,
  title={Curriculum learning for ab initio deep learned refractive optics},
  author={Yang, Xinge and Fu, Qiang and Heidrich, Wolfgang},
  journal={Nature Communications},
  volume={15},
  number={1},
  pages={6572},
  year={2024},
  publisher={Nature Publishing Group UK London}
}

@article{cote2019extrapolating,
  title={Extrapolating from lens design databases using deep learning},
  author={C{\^o}t{\'e}, Geoffroi and Lalonde, Jean-Fran{\c{c}}ois and Thibault, Simon},
  journal={Optics Express},
  volume={27},
  number={20},
  pages={28279--28292},
  year={2019},
  publisher={Optica Publishing Group}
}

@article{cote2021deep,
  title={Deep learning-enabled framework for automatic lens design starting point generation},
  author={C{\^o}t{\'e}, Geoffroi and Lalonde, Jean-Fran{\c{c}}ois and Thibault, Simon},
  journal={Optics Express},
  volume={29},
  number={3},
  pages={3841--3854},
  year={2021},
  publisher={Optica Publishing Group}
}

@article{gao2025exploring,
  title={Exploring Quasi-Global Solutions to Compound Lens Based Computational Imaging Systems},
  author={Gao, Yao and Jiang, Qi and Gao, Shaohua and Sun, Lei and Yang, Kailun and Wang, Kaiwei},
  journal={IEEE Transactions on Computational Imaging},
  year={2025},
  publisher={IEEE}
}

@article{yang2017automated,
  title={Automated design of freeform imaging systems},
  author={Yang, Tong and Jin, Guo-Fan and Zhu, Jun},
  journal={Light: Science \& Applications},
  volume={6},
  number={10},
  pages={e17081--e17081},
  year={2017},
  publisher={Nature Publishing Group}
}

@article{zhang2021towards,
  title={Towards automatic freeform optics design: coarse and fine search of the three-mirror solution space},
  author={Zhang, Benqi and Jin, Guofan and Zhu, Jun},
  journal={Light: Science \& Applications},
  volume={10},
  number={1},
  pages={65},
  year={2021},
  publisher={Nature Publishing Group UK London}
}

@article{zhang2023large,
  title={Large depth-of-field ultra-compact microscope by progressive optimization and deep learning},
  author={Zhang, Yuanlong and Song, Xiaofei and Xie, Jiachen and Hu, Jing and Chen, Jiawei and Li, Xiang and Zhang, Haiyu and Zhou, Qiqun and Yuan, Lekang and Kong, Chui and others},
  journal={Nature Communications},
  volume={14},
  number={1},
  pages={4118},
  year={2023},
  publisher={Nature Publishing Group UK London}
}

@article{gissibl2016sub,
  title={Sub-micrometre accurate free-form optics by three-dimensional printing on single-mode fibres},
  author={Gissibl, Timo and Thiele, Simon and Herkommer, Alois and Giessen, Harald},
  journal={Nature communications},
  volume={7},
  number={1},
  pages={11763},
  year={2016},
  publisher={Nature Publishing Group UK London}
}

@article{wang2015witnessing,
  title={Witnessing magnetic twist with high-resolution observation from the 1.6-m New Solar Telescope},
  author={Wang, Haimin and Cao, Wenda and Liu, Chang and Xu, Yan and Liu, Rui and Zeng, Zhicheng and Chae, Jongchul and Ji, Haisheng},
  journal={Nature Communications},
  volume={6},
  number={1},
  pages={7008},
  year={2015},
  publisher={Nature Publishing Group UK London}
}

@article{liu2025global,
  title={Global information selectively guided gradient descent for ab initio optical design},
  author={Liu, Xiaobing and Zhang, Xingxiang and Fu, Tianjiao and Wang, Kaizhi and Sun, Fukun and Bai, Tongzheng and Wang, Duo},
  journal={Optics \& Laser Technology},
  volume={184},
  pages={112497},
  year={2025},
  publisher={Elsevier}
}

@article{gao2025neuro,
  title={Neuro-inspired automated lens design},
  author={Gao, Yao and Sun, Lei and Gao, Shaohua and Jiang, Qi and Yang, Kailun and Hu, Weijian and Qian, Xiaolong and Li, Wenyong and Van Gool, Luc and Wang, Kaiwei},
  journal={arXiv preprint arXiv:2510.09979},
  year={2025}
}

@article{liu2024deepseek,
  title={Deepseek-v3 technical report},
  author={Liu, Aixin and Feng, Bei and Xue, Bing and Wang, Bingxuan and Wu, Bochao and Lu, Chengda and Zhao, Chenggang and Deng, Chengqi and Zhang, Chenyu and Ruan, Chong and others},
  journal={arXiv preprint arXiv:2412.19437},
  year={2024}
}

@article{yang2025qwen3,
  title={Qwen3 technical report},
  author={Yang, An and Li, Anfeng and Yang, Baosong and Zhang, Beichen and Hui, Binyuan and Zheng, Bo and Yu, Bowen and Gao, Chang and Huang, Chengen and Lv, Chenxu and others},
  journal={arXiv preprint arXiv:2505.09388},
  year={2025}
}

@article{guo2025deepseek,
  title={Deepseek-r1: Incentivizing reasoning capability in llms via reinforcement learning},
  author={Guo, Daya and Yang, Dejian and Zhang, Haowei and Song, Junxiao and Zhang, Ruoyu and Xu, Runxin and Zhu, Qihao and Ma, Shirong and Wang, Peiyi and Bi, Xiao and others},
  journal={arXiv preprint arXiv:2501.12948},
  year={2025}
}

@inproceedings{zhang2023xuanyuan,
  title={Xuanyuan 2.0: A large chinese financial chat model with hundreds of billions parameters},
  author={Zhang, Xuanyu and Yang, Qing},
  booktitle={Proceedings of the 32nd ACM international conference on information and knowledge management},
  pages={4435--4439},
  year={2023}
}

@inproceedings{lai2025llmlight,
  title={Llmlight: Large language models as traffic signal control agents},
  author={Lai, Siqi and Xu, Zhao and Zhang, Weijia and Liu, Hao and Xiong, Hui},
  booktitle={Proceedings of the 31st ACM SIGKDD Conference on Knowledge Discovery and Data Mining V. 1},
  pages={2335--2346},
  year={2025}
}

@misc{cursor2024,
  title        = {Cursor: AI-powered Code Editor},
  author={Cursor},
  howpublished = {\url{https://www.cursor.com/en}},
}

@inproceedings{besta2024graph,
  title={Graph of thoughts: Solving elaborate problems with large language models},
  author={Besta, Maciej and Blach, Nils and Kubicek, Ales and Gerstenberger, Robert and Podstawski, Michal and Gianinazzi, Lukas and Gajda, Joanna and Lehmann, Tomasz and Niewiadomski, Hubert and Nyczyk, Piotr and others},
  booktitle={Proceedings of the AAAI Conference on Artificial Intelligence},
  volume={38},
  number={16},
  pages={17682--17690},
  year={2024}
}

@misc{qian2025toolrl,
      title={ToolRL: Reward is All Tool Learning Needs}, 
      author={Cheng Qian and Emre Can Acikgoz and Qi He and Hongru Wang and Xiusi Chen and Dilek Hakkani-Tür and Gokhan Tur and Heng Ji},
      year={2025},
      eprint={2504.13958},
      archivePrefix={arXiv},
      primaryClass={cs.LG},
      url={https://arxiv.org/abs/2504.13958}, 
}

@article{singh2025openai,
  title={Openai gpt-5 system card},
  author={Singh, Aaditya and Fry, Adam and Perelman, Adam and Tart, Adam and Ganesh, Adi and El-Kishky, Ahmed and McLaughlin, Aidan and Low, Aiden and Ostrow, AJ and Ananthram, Akhila and others},
  journal={arXiv preprint arXiv:2601.03267},
  year={2025}
}

@misc{liu2025understandingr1zeroliketrainingcritical,
      title={Understanding R1-Zero-Like Training: A Critical Perspective}, 
      author={Zichen Liu and Changyu Chen and Wenjun Li and Penghui Qi and Tianyu Pang and Chao Du and Wee Sun Lee and Min Lin},
      year={2025},
      eprint={2503.20783},
      archivePrefix={arXiv},
      primaryClass={cs.LG},
      url={https://arxiv.org/abs/2503.20783}, 
}

@article{gao2024design,
  title={Design, analysis, and manufacturing of a glass-plastic hybrid minimalist aspheric panoramic annular lens},
  author={Gao, Shaohua and Jiang, Qi and Liao, Yiqi and Qiu, Yi and Ying, Wanglei and Yang, Kailun and Wang, Kaiwei and Zhang, Benhao and Bai, Jian},
  journal={Optics \& Laser Technology},
  volume={177},
  pages={111119},
  year={2024},
  publisher={Elsevier}
}

@inproceedings{dong2024survey,
  title={A survey on in-context learning},
  author={Dong, Qingxiu and Li, Lei and Dai, Damai and Zheng, Ce and Ma, Jingyuan and Li, Rui and Xia, Heming and Xu, Jingjing and Wu, Zhiyong and Chang, Baobao and others},
  booktitle={Proceedings of the 2024 conference on empirical methods in natural language processing},
  pages={1107--1128},
  year={2024}
}

@article{touvron2023llama,
  title={Llama: Open and efficient foundation language models},
  author={Touvron, Hugo and Lavril, Thibaut and Izacard, Gautier and Martinet, Xavier and Lachaux, Marie-Anne and Lacroix, Timoth{\'e}e and Rozi{\`e}re, Baptiste and Goyal, Naman and Hambro, Eric and Azhar, Faisal and others},
  journal={arXiv preprint arXiv:2302.13971},
  year={2023}
}

@article{rafailov2023direct,
  title={Direct preference optimization: Your language model is secretly a reward model},
  author={Rafailov, Rafael and Sharma, Archit and Mitchell, Eric and Manning, Christopher D and Ermon, Stefano and Finn, Chelsea},
  journal={Advances in neural information processing systems},
  volume={36},
  pages={53728--53741},
  year={2023}
}

@inproceedings{devlin2019bert,
  title={Bert: Pre-training of deep bidirectional transformers for language understanding},
  author={Devlin, Jacob and Chang, Ming-Wei and Lee, Kenton and Toutanova, Kristina},
  booktitle={Proceedings of the 2019 conference of the North American chapter of the association for computational linguistics: human language technologies, volume 1 (long and short papers)},
  pages={4171--4186},
  year={2019}
}

@article{chowdhery2023palm,
  title={Palm: Scaling language modeling with pathways},
  author={Chowdhery, Aakanksha and Narang, Sharan and Devlin, Jacob and Bosma, Maarten and Mishra, Gaurav and Roberts, Adam and Barham, Paul and Chung, Hyung Won and Sutton, Charles and Gehrmann, Sebastian and others},
  journal={Journal of Machine Learning Research},
  volume={24},
  number={240},
  pages={1--113},
  year={2023}
}

@inproceedings{lewis2020bart,
  title={BART: Denoising sequence-to-sequence pre-training for natural language generation, translation, and comprehension},
  author={Lewis, Mike and Liu, Yinhan and Goyal, Naman and Ghazvininejad, Marjan and Mohamed, Abdelrahman and Levy, Omer and Stoyanov, Veselin and Zettlemoyer, Luke},
  booktitle={Proceedings of the 58th annual meeting of the association for computational linguistics},
  pages={7871--7880},
  year={2020}
}

@article{nakano2021webgpt,
  title={Webgpt: Browser-assisted question-answering with human feedback},
  author={Nakano, Reiichiro and Hilton, Jacob and Balaji, Suchir and Wu, Jeff and Ouyang, Long and Kim, Christina and Hesse, Christopher and Jain, Shantanu and Kosaraju, Vineet and Saunders, William and others},
  journal={arXiv preprint arXiv:2112.09332},
  year={2021}
}

@misc{openaigpto1,
  title={Introducing OpenAI o1-preview},
  author={OpenAI},
  howpublished={\url{https://openai.com/index/introducing-openai-o1-preview/}},
  year={2024}
}

@misc{openaigpt4o,
  title={Hello GPT-4o},
  author={OpenAI},
  howpublished={\url{https://openai.com/index/hello-gpt-4o/}},
  year={2024}
}

@misc{anthropic2024claude,
  title={Claude 3.5 Sonnet},
  author={Anthropic},
  howpublished={\url{https://www.anthropic.com/news/claude-3-5-sonnet}},
  year={2024}
}

@misc{deepseekai2025deepseekr1,
      title={DeepSeek-R1: Incentivizing Reasoning Capability in LLMs via Reinforcement Learning}, 
      author={DeepSeek-AI},
      year={2025},
      eprint={2501.12948},
      archivePrefix={arXiv},
      primaryClass={cs.CL},
      url={https://arxiv.org/abs/2501.12948}, 
}
\bibliographystyle{icml2025}

\newpage
\appendix
\onecolumn

\section{Fundamentals of Geometrical Optics}
\label{supp:fundamentals}

This section provides the essential background of geometrical optics required to understand the parameterization and physical constraints used in \ourmethod.

\subsection{Basic Components of an Optical System}
A fundamental imaging system in geometrical optics consists of several key planes and elements that govern the propagation of light:

\begin{itemize}
    \item \textbf{Object Plane (OBJ):} The plane where the source or the scene being imaged is located. Light rays originate from points on this plane.
    \item \textbf{Aperture Stop (STOP):} A physical opening (often an iris) that limits the amount of light reaching the image plane. It determines the system's \textit{F-number} and significantly influences image aberrations.
    \item \textbf{Optical Elements:} A series of refractive surfaces (lenses) that redirect light rays. In our \ourmethod, these are defined by the radii and thicknesses in the prescription.
    \item \textbf{Image Plane (IMAGE):} The plane where light rays converge to form a reproduction of the object. For a system to be in focus, the rays originating from a single point on the object plane must ideally converge to a single point on the image plane.
\end{itemize}

\textbf{The Concept of Imaging:} 
Geometrically, imaging is the process of mapping points from the object space to corresponding points in the image space. A feasible design must ensure that the light rays emitted from the object are collected by the optical elements and transformed into a high-quality (sharp and undistorted) representation at the image plane. The goal of \ourmethod is to optimize the parameters of the optical elements to satisfy this mapping while adhering to physical constraints.

\subsection{The Optical Prescription}
In optical engineering, a lens system is defined by a \textit{prescription table}, which lists the structural parameters for each surface. Our \textbf{Optical Prescription Completion} task focuses on the following numerical values:
\begin{itemize}
    \item \textbf{Curvature Radius ($R$):} The radius of the spherical surface. It determines the optical power; a smaller radius implies a more "curved" surface and stronger light-bending capability.
    \item \textbf{Thickness ($d$):} The axial distance between consecutive surfaces. This includes both the lens material (center thickness) and the air gaps between elements.
    \item \textbf{Refractive Index:} A material property representing the ratio of the speed of light in a vacuum to that in the medium.
    \item \textbf{Abbe Number:} A measure of the material's dispersion, characterizing how the refractive index varies with wavelength. A higher Abbe number indicates lower dispersion, which is essential for correcting chromatic aberrations.
\end{itemize}

\subsection{First-Order System Properties}
We provide technical definitions for the optical design parameters and evaluation metrics used throughout this work.

\begin{itemize}
    \item \textbf{Effective Focal Length (EFFL):} The distance between the principal plane and the focal point. It determines the system's magnification and is the primary factor in defining the lens's scale.
    \item \textbf{F-number ($F/\#$):} The ratio of the focal length to the diameter of the entrance pupil. It governs the light-gathering power and the theoretical resolution limit of the system.
    \item \textbf{Field of View (FOV):} The maximum angular range of the object space that the system can capture, typically measured across the diagonal of the image sensor.
    \item \textbf{Back Focal Length (BFL):} The physical distance from the vertex of the last optical surface to the paraxial image plane. Maintaining an adequate BFL is crucial for accommodating mechanical components such as filters, shutters, or sensor housings.
    \item \textbf{Root Mean Square (RMS) Spot Size:} The root mean square of the distances between the intersection points of rays with the image plane and their centroid. It provides a statistical measure of the image blur. A smaller RMS spot size indicates superior convergence of light and higher image sharpness.
\end{itemize}

\subsection{Physical Constraints}
Detailed explanations for physical constraints in optical design.

\begin{itemize}
    \item \textbf{Positive Thickness:} In a physical assembly, the center thickness ($d_c$) must be positive. 
    \item \textbf{Edge Thickness ($d_e$):} The thickness at the clear aperture of the lens. Lenses with negative or near-zero edge thickness (known as "knife-edges") are impossible to manufacture or mount.
\end{itemize}

All physical plausibility constraints are integrated into the Structure Reward in our proposed Optical Lexicographic Reward function.

\subsection{Imaging process}



\begin{figure}[htbp]
  \centering
  \begin{minipage}[c]{0.6\linewidth}
    \centering
    \includegraphics[width=\linewidth]{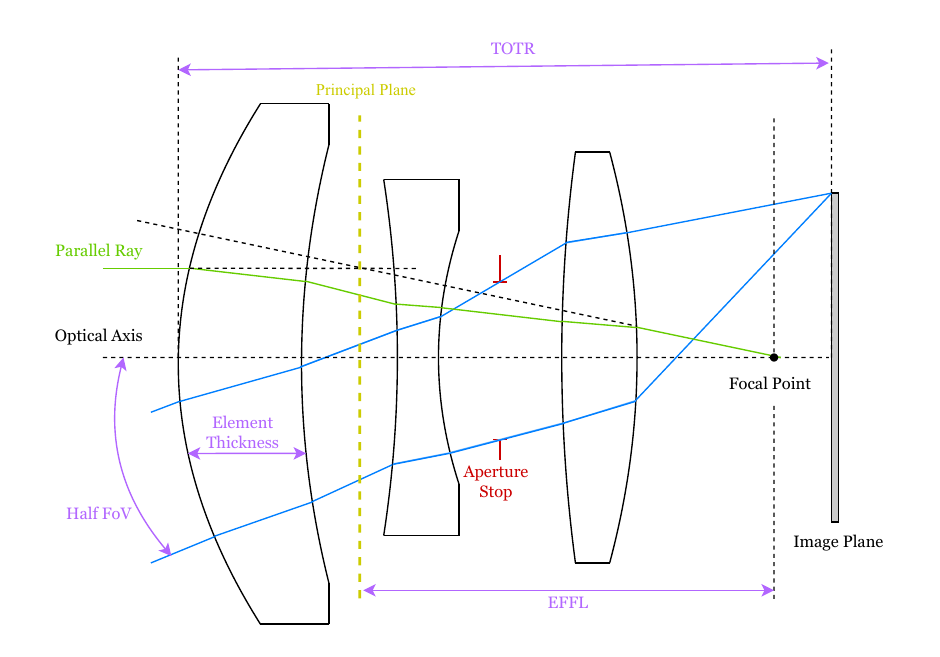}
    \caption{Illustration of an imaging optical system.}
    \label{fig:optic_system}
  \end{minipage}
  \hfill
  \begin{minipage}[c]{0.38\linewidth}
    \centering
    \resizebox{\linewidth}{!}{
      \begin{tabular}{ccccc}
        \toprule
        \textbf{Surface} & \textbf{\makecell{Radius \\ (mm)}} & \textbf{\makecell{Thickness \\ (mm)}} & \textbf{Material} & \textbf{\makecell{Semi-Diameter \\ (mm)}} \\
        \midrule
        \textbf{OBJ} & $\infty$ & $\infty$ &        & $\infty$ \\
        \textbf{1}  & 40.940   & 8.740    & N-LAK7 & 23.000 \\
        \textbf{2}  & $\infty$ & 11.050   &        & 23.000 \\
        \textbf{3}  & -55.650  & 2.780    & N-SF8  & 20.000 \\
        \textbf{4}  & 39.750   & 6.630    &        & 20.000 \\
        \textbf{STO}& $\infty$ & 1.000    &        & 14.432 \\
        \textbf{6}  & 107.560  & 9.540    & N-LAK22 & 22.000 \\
        \textbf{7}  & -43.330  & 73.587   &        & 22.000 \\
        \textbf{IMA}& $\infty$ & -        &        & 16.803 \\
        \bottomrule
      \end{tabular}
    }
    \captionof{table}{Lens data parameters for the optical system.}
    \label{tab:lens_data}
  \end{minipage}
\end{figure}

The diagram Fig.~\ref{fig:optic_system} illustrates a typical multi-element coaxial optical system, where the fundamental imaging principle involves utilizing a combination of lenses with varying curvatures and refractive indices to precisely map light field information from the object space to the sensor plane in the image space after multiple refraction corrections. Tab.~\ref{tab:lens_data} shows the corresponding lens data parameters. Physically, the entire system is strictly rotationally symmetric about the Optical Axis, which serves as the reference zero for defining all heights and angles. Light rays enter from the left, traverse through the lens group containing specific Element Thicknesses, and finally reach the Image Plane. The axial total distance from the front surface of the first lens to the image plane is defined as the TOTR (TOtal TRack length), which is a critical mechanical metric for evaluating the compactness of optical modules, particularly in mobile phones or compact cameras.

Regarding the definition of core optical parameters, Gaussian optics introduces the Principal Plane as an equivalent refraction plane to simplify the complex continuous refraction process of multiple lenses. The green Parallel Ray shown in the figure represents on-axis signals from infinity; after being refracted by the system, these rays converge at the Focal Point on the optical axis. The distance from the principal plane to the focal point is the EFFL (Effective Focal Length), which directly determines the optical system's magnification and image size. It is crucial to note that the EFFL is an optical parameter calculated based on the principal plane and is often not equivalent to the mechanical distance from the surface of the last lens element to the image plane (known as the Back Focal Length). This distinction is key to understanding the difference between the thick lens model and the pinhole model.

The field of view and energy distribution are primarily governed by the field angle and the aperture stop. The Field of View (FoV) represents the total angular extent that the optical system can perceive, which effectively defines the overall coverage of the imaging area. As the marginal rays originating from the edge of this field enter the system, they are regulated by the aperture stop, which is highlighted by the red line in the diagram. Functioning as the system's ``pupil'', the aperture stop constrains the beam cross-section to determine both the luminous flux (F-number) and the depth of field. Furthermore, it performs the critical role of filtering out peripheral rays that would otherwise induce significant aberrations, such as spherical aberration or coma, thereby ensuring that the image formed on the focal plane achieves optimal optical quality and sharpness.


\section{More Details about Optical Lexicographic Reward}






\subsection{Structural Integrity Verification (Format)}

Before evaluating any optical properties, the generated design must strictly adhere to the Optical Data Description Language (ODDL) format. We implement a rule-based parsing engine to enforce topological validity and data completeness.

\subsubsection{Verification Pipeline}

The format check $S_{\text{fmt}}$ is a binary-like gatekeeper ($S_{\text{fmt}} \in \{0, 1\}$) that ensures the output is physically interpretable. The verification proceeds in four hierarchical stages:

\begin{enumerate}
    \item \textbf{Basic Existence:} The output must contain essential system specifications (e.g., Effective Focal Length, EFFL) and a valid surface data table.

    \item \textbf{Topological Constraints:} 
    \begin{itemize}
        \item The first surface (Index 0) must be defined as the \textbf{Object (OBJ)} plane.
        \item The last surface must be explicitly marked as the \textbf{Image (IMA)} plane.
        \item A unique \textbf{Aperture Stop (STOP)} must be defined, ensuring it is strictly located between the object and image planes.
    \end{itemize}

    \item \textbf{Data Completeness:} Every optical surface must include numerical values for Curvature Radius ($R$), Thickness ($T$), and a valid Material identifier ($N_d$).

    \item \textbf{Physical Causality:} To prevent "hallucinated" physics (e.g., embedded image planes), we enforce boundary conditions:
    \begin{equation}
        \text{Mat}_{N-1} \in \{\varnothing, \text{AIR}, \text{VAC}\} \quad \land \quad T_{N-1} > 0
    \end{equation}
    This requires the medium preceding the image plane to be air/vacuum, and the Back Focal Length (BFL) to be strictly positive.
\end{enumerate}

\subsubsection{Scoring Policy}

If any violation is detected at any stage, the format score $S_{\text{fmt}}$ is set to 0, and the entire evaluation terminates immediately to penalize invalid structures. Otherwise, $S_{\text{fmt}} = 1$.

\subsection{Physics-based Reward Implementation}

The physics verification module assesses the fundamental optical properties using first-order paraxial ray tracing. The reward function $S_{\text{phy}}$ aims to guide the agent towards generating designs that strictly compliant with the target effective focal length (EFFL) and maintain correct image plane positioning.

\subsubsection{Paraxial Ray Tracing}

We implement a differentiable paraxial ray tracing engine based on the $y-u$ method. For a system with $k$ surfaces, the ray height $y$ and angle $u$ at surface $i$ are propagated as:

\begin{align}
    \phi_i &= (n'_i - n_i) c_i \\  
    u'_i &= \frac{n_i u_i - y_i \phi_i}{n'_i} \\ 
    y_{i+1} &= y_i + u'_i t_i 
\end{align}
where $n, n'$ are refractive indices before and after the surface, $c$ is the curvature, and $t$ is the thickness. By tracing a marginal ray parallel to the optical axis ($u_0=0, y_0=h$), we derive the calculated EFFL ($f_{\text{calc}}$) and the residual ray height at the image plane ($y_{\text{img}}$).

\subsubsection{Scoring Function}

The physics reward $S_{\text{phy}}$ is a weighted sum of the focal length accuracy score ($S_{\text{f}}$) and the convergence score ($S_{\text{c}}$):
\begin{equation}
    S_{\text{phy}} = w_f \cdot S_{\text{f}} + w_c \cdot S_{\text{c}}
\end{equation}
where $w_f = 0.6$ and $w_c = 0.4$.

\textbf{1. Focal Length Accuracy ($S_{\text{f}}$):} To balance coarse exploration and fine-tuning, we employ a mixed exponential decay function based on the relative error $\epsilon = |f_{\text{calc}} - f_{\text{target}}| / f_{\text{target}}$:
\begin{equation}
    S_{\text{f}} = 0.7 \cdot \exp\left(-\frac{\epsilon}{\alpha_1}\right) + 0.3 \cdot \exp\left(-\frac{\epsilon}{\alpha_2}\right)
\end{equation}
Here, $\alpha_1 = 0.02$ serves as a strict sensitivity parameter (rewarding $<2\%$ error), while $\alpha_2 = 0.10$ serves as a loose parameter (guiding agents within $10\%$ error).

\textbf{2. Ray Convergence ($S_{\text{c}}$):} This component ensures the rays focus precisely on the defined image plane. It is calculated using the residual ray height $y_{\text{img}}$:
\begin{equation}
    S_{\text{c}} = \exp\left(-\frac{|y_{\text{img}}|}{\beta}\right)
\end{equation}
where $\beta = 1.0$ mm is the scaling factor.

\subsubsection{Gating Threshold}

A binary gating indicator $\delta_{\text{pass}}$ controls the activation of subsequent reward modules (e.g., spot size optimization). It is defined as:
\begin{equation}
    \delta_{\text{pass}} = \mathbb{I}(\epsilon < 0.05 \land |y_{\text{img}}| < 0.1 \text{ mm})
\end{equation}
where $\mathbb{I}(\cdot)$ is the indicator function. This ensures that computationally expensive high-order optimization is only performed on valid first-order designs.
\subsection{Image Quality Optimization (RMS)}

we employ a multi-ray paraxial sampling strategy as a computationally efficient proxy to evaluate focusing consistency and field flatness.

\subsubsection{Beam Sampling Strategy}

We explicitly formulate the root-mean-square (RMS) spot radius by sampling rays across the pupil diameter and field of view (FOV).

For a given field angle $\theta_k \in \{0, \text{FOV}/2\}$, we trace a discrete set of rays originating from normalized pupil coordinates $\rho \in \{-1, -0.5, 0, 0.5, 1\}$. The hit position $h_{k, \rho}$ of each ray on the image plane is computed. The RMS spot radius for field $k$ is defined as:
\begin{equation}
    \sigma_k = \sqrt{\frac{1}{N} \sum_{\rho} \left( h_{k, \rho} - \bar{h}_k \right)^2 }
\end{equation}
where $\bar{h}_k$ is the centroid of the spot.

\subsubsection{Reward Formulation}

To enforce uniform image quality across the entire field, we optimize for the worst-case scenario. The final metric is the maximum RMS error across all sampled fields:
\begin{equation}
    \sigma_{\text{max}} = \max_{k} (\sigma_k)
\end{equation}

The RMS reward score $S_{\text{rms}}$ is then determined by a dynamic exponential decay function:
\begin{equation}
    S_{\text{rms}} = \exp\left( -\frac{\sigma_{\text{max}}}{\gamma} \right)
\end{equation}
Here, the scaling factor $\gamma = \max(0.05, 0.01 \cdot f_{\text{effl}})$ adapts to the system's scale, tolerating a spot size proportional to roughly 1\% of the effective focal length.

\section{Training with DrGRPO}
\label{subsec:DrGRPO}

\ourmethod is trained using the DrGRPO algorithm~\cite{liu2025understandingr1zeroliketrainingcritical}, an unbiased variant of Group Relative Policy Optimization designed to mitigate length bias and optimization instability. For each user demand $q$, a group of $G$ lens designs $\{o_1, o_2, \dots, o_G\}$ is sampled from the current policy $\pi_{\theta_{\text{old}}}$. 

Unlike the standard GRPO formulation, DrGRPO simplifies the advantage estimation by removing the standard deviation normalization to prevent gradient instability when reward variance is low. The advantage $A_i$ for the $i$-th output is computed solely by centering the rewards:
\begin{equation}
A_i = r_i - \frac{1}{G}\sum_{j=1}^G r_j.
\end{equation}

Furthermore, to avoid the optimization bias that encourages unnecessarily long generations, DrGRPO aggregates the token-level objectives without normalizing by the specific sequence length. The policy is optimized by maximizing the following objective:
\begin{equation}
J_{\text{DrGRPO}}(\theta) = \mathbb{E}_{q \sim \mathcal{D}, \{o_i\}_{i=1}^G \sim \pi_{\theta_{\text{old}}}} \left[ \frac{1}{G} \sum_{i=1}^G \sum_{t=1}^{|o_i|} \left( \min \left( \rho_{i,t} A_i, \operatorname{clip}\left( \rho_{i,t}, 1-\varepsilon, 1+\varepsilon \right) A_i \right) - \beta D_{\text{KL}}(\pi_\theta || \pi_{\text{ref}}) \right) \right]
\end{equation}
where $\rho_{i,t} = \frac{\pi_\theta(o_{i,t}|q, o_{i,<t})}{\pi_{\text{old}}(o_{i,t}|q, o_{i,<t})}$ is the token-level probability ratio. By eschewing length normalization and standard deviation scaling, this formulation ensures that the model optimizes for efficiency and correctness rather than exploiting length-based reward loopholes.


\end{document}